%% file: main.tex
\def\year{2021}\relax
\documentclass[letterpaper]{article} % DO NOT CHANGE THIS
\usepackage{aaai21}  % DO NOT CHANGE THIS
\usepackage{times}  % DO NOT CHANGE THIS
\usepackage{helvet} % DO NOT CHANGE THIS
\usepackage{courier}  % DO NOT CHANGE THIS
\usepackage[hyphens]{url}  % DO NOT CHANGE THIS
\usepackage{graphicx} % DO NOT CHANGE THIS
\usepackage{natbib}  % DO NOT CHANGE THIS AND DO NOT ADD ANY OPTIONS TO IT
\usepackage{caption} % DO NOT CHANGE THIS AND DO NOT ADD ANY OPTIONS TO IT
\usepackage{amsmath}
\usepackage{amssymb}
\input{math_commands.tex}

\usepackage{xcolor}
\usepackage{multirow}
\usepackage[switch]{lineno}

\newcommand{\cut}[1]{}

\title{A Fully Tensorized Recurrent Neural Network}
\author{
    Charles C. Onu,\textsuperscript{\rm 1,2} Jacob E. Miller,\textsuperscript{\rm 3} Doina Precup\textsuperscript{\rm 1,4} \\ 
}
\affiliations{
    \textsuperscript{\rm 1}Mila - McGill University\\ 
    \textsuperscript{\rm 2}Ubenwa Health \\
    \textsuperscript{\rm 3}Mila - Université de Montréal \\
    \textsuperscript{\rm 4} Google DeepMind
}
\begin{document}
% \linenumbers

\maketitle

\begin{abstract}

Recurrent neural networks (RNNs) are powerful tools for sequential modeling, but typically require significant overparameterization and regularization to achieve optimal performance. This leads to difficulties in the deployment of large RNNs in resource-limited settings, while also introducing complications in hyperparameter selection and training. To address these issues, we introduce a ``fully tensorized'' RNN architecture which jointly encodes the separate weight matrices within each recurrent cell using a lightweight tensor-train (TT) factorization. This approach represents a novel form of weight sharing which reduces model size by several orders of magnitude, while still maintaining similar or better performance compared to standard RNNs. Experiments on image classification and speaker verification tasks demonstrate further benefits for reducing inference times and stabilizing model training and hyperparameter selection.

\end{abstract}

\section{Introduction}

Recurrent neural networks (RNNs) represent a model family that is well-suited for tasks involving sequential data. Although early RNNs were limited by the problem of vanishing gradients during training, this was largely solved by the development of gated RNNs such as long short-term memory (LSTM) and gated recurrent unit (GRU) models~\cite{hochreiter1997,cho2014}, which employ a collection of independent weight matrices to control the propagation of gradients. Such models have allowed RNNs to attain impressive performance in tasks such as speech recognition, language modeling, time series forecasting, and video classification.

RNNs typically employ large hidden states to achieve better performance in difficult modeling tasks, which in turn leads to a significant increase in parameters used to specify large weight matrices. The memory and compute issues associated with running such models, particularly in the limited setting of mobile and embedded devices, has led to the use of various techniques for model compression, including model distillation~\cite{hinton2015}, alternate matrix decompositions~\cite{sainath2013}, and quantization of network weights~\cite{he2019}. The use of such compression strategies is supported by the observation that standard representations of neural networks contain significant amounts of redundancy~\cite{denil2013,cheng2015}.

In this work we use the tensor-train (TT) formalism, a means of efficiently representing multi-modal tensors, to achieve significant compression of the model parameters associated with various RNN architectures. In contrast to previous work~\cite{tjandra2017,yang2017tensor}, we apply the TT formalism jointly to all weight matrices within the RNN, leading to a ``fully tensorized'' form of weight sharing, where various gate matrices are encoded within a single TT format. This permits the development of extremely lightweight, end-to-end trainable models, even in the presence of high-dimensional hidden states or input representations. 

Experiments on image classification and speaker verification show that fully tensorized TT-RNNs give comparable or better performance relative to their uncompressed counterparts. We demonstrate that our method leads to state-of-the-art performance on the LibriSpeech dataset, producing a 16\% reduction in speaker verification error while simultaneously allowing for a 200-fold compression in model parameters.

% INTRO SECTION TO BE COMPLETED

%  - this work is concerned with model compression in recurrent neural networks, precisely reduction in no of params of rnn. Smaller models implies lower-memory usage, require lesser disk space and suitability to edge devices among other benefits.
 
%  - the fact that DNNs are over parameterised is no secret. Can predict weights (dinh 2014). several proposed methods in the literature are dependent on this: model distillation, quantisation.
 
%  In this work, we build off the assumption that there exists a low-rank representation of the set of weight matrices in a typical RNN, which is sufficient to capture an equivalent fully-rank matrix. in particular, we apply technique of tensor-train decomposition to restructure the functions for the gates and memory cell of an LSTM. can also be applied to GRU or other cells with negligible modifications.
 
%  (characteristics of model) This results to a model that is...1000 compression factor, yet.
 
% -  We carefully study the behaviour and performance of this architecture experimentally on sequential character prediction (using the MNIST dataset). We further test the performance of the architecture on the more complicated speech task of speaker verification. 
 
%  Results to confirm..

\subsection{Notation}
We use bold lower-case letters $\va$ to denote vectors, bold upper-case letters $\mW$ to denote matrices, and bold calligraphic letters to denote tensors $\tT$. Tensor elements are indexed as $\va(i)$, $\mW(i,j)$ and $\tT(i_1, i_2,..,i_d)$, for the respective cases of vectors, matrices, and more general $d$'th-order tensors. The notation $\va * \vb$ represents the element-wise Hadamard product between vectors of equal size. The collection of integers $\{1, 2, \ldots, n\}$ is denoted as $[n]$.

\section{Recurrent Neural Networks and the Tensor-Train Decomposition}
In this section we first give an overview of common RNN architectures, before introducing the tensor-train decomposition and describing its use for ``tensorizing'' large weight matrices within neural networks. %This new representation is extremely frugal in terms of model size on disk, memory usage, and runtime, while still remaining expressive and scaling favorably with increasing hidden state size compared with traditional RNNs.

\subsection{RNN Architectures}
Recurrent neural networks (RNN) define a paradigm for learning from sequential data. The recurrent unit of an RNN defines an iterative procedure whose outputs and hidden state at each time step $t$ are a non-linear function of $\vx_{t}$, the input at $t$, and $\vh_{t-1}$, the hidden state at time $t-1$. Many different functions have been proposed for this nonlinear recurrent unit, and we describe two representative choices, long short-term memory (LSTM) and gated recurrent unit (GRU).

% \subsubsection{Simple RNN}
% The hidden state at time $t$ of the computational unit of a simple RNN is given by:

% \begin{equation}
%     \begin{split}
%     \vy_{t} &= \varphi_y(\mW^{(y)} \vh_{t} + \vb^{(y)}) \\
%     \vh_{t} &= \varphi_h(\mW^{(h)} \vx_{t} + \mU^{(h)} \vh_{t-1} + \vb^{(h)})
%     \end{split}
% \end{equation}
% where $\vh_{t} \in \R^D$ is the $D$-dimensional hidden state and $\vx_{t} \in \R^M$ is the input. The weight parameters consist of the matrices and bias vectors $\mW^{(y)} \in \R^{N \times D}$, $\vb^{(y)} \in \R^N$, $\mW^{(h)} \in \R^{D \times M}$, $\mU^{(h)} \in \R^{D \times D}$, and $\vb^{(h)} \in \R^D$, with $\varphi_y$ and $\varphi_h$ denoting activation functions such as sigmoid or tanh.

% This formulation suffers from the vanishing or exploding gradients problem when long sequence lengths are used, which spurred the design of more sophisticated RNN units employing multiplicative gating mechanisms.

\subsubsection{Long Short-Term Memory}
The LSTM cell uses three ``gates'' to control the flow of information, and divides its hidden state into a memory cell state $\vc$ and regular hidden state $\vh$, of identical dimension $D$. These are jointly updated as
\begin{equation}
\begin{split}
    \vc_{t} &= \vu_{t} * \Tilde{\vc}_{t} + \vf_{t} * \vc_{t-1} \\
    \vh_{t} &= \vo_{t} * \tanh(\vc_{t}),
\end{split}
\end{equation}

\noindent where the candidate cell state ($\Tilde{\vc}_{t}$), update gate ($\vu_{t}$), forget gate ($\vf_{t}$), and output gate ($\vo_{t}$) vectors are given by
\begin{equation}
\begin{split}
\label{eq:lstm_gates}
    \Tilde{\vc}_{t} &= \tanh(\mW^{(c)} \vx_{t} + \mU^{(c)} \vh_{t-1} + \vb^{(c)})  \\
    \vu_{t} &= \sigma(\mW^{(u)} \vx_{t} + \mU^{(u)} \vh_{t-1} + \vb^{(u)}) \\
    \vf_{t} &= \sigma(\mW^{(f)} \vx_{t} + \mU^{(f)} \vh_{t-1} + \vb^{(f)}) \\
    \vo_{t} &= \sigma(\mW^{(o)} \vx_{t} + \mU^{(o)} \vh_{t-1} + \vb^{(o)}).
\end{split}
\end{equation}

In the above, $\vx_{t} \in \mathbb{R}^M$ and $\vh_{t} \in \mathbb{R}^D$ are the input and hidden state vectors respectively, while  $\mW^{(c)}, \mW^{(u)}, \mW^{(f)}, \mW^{(o)} \in \mathbb{R}^{D \times M}$ are the input-hidden transition matrices, and $\mU^{(c)}, \mU^{(u)}, \mU^{(f)}, \mU^{(o)} \in \mathbb{R}^{D \times D}$ are the hidden-hidden transition matrices.

\subsubsection{Gated Recurrent Unit}
The GRU is defined by two (update and relevance) gates and a single hidden state
\begin{equation}
    \vh_{t} = \vu_{t} * \Tilde{\vh}_{t} + (1 - \vu_{t}) * \vh_{t-1},
\end{equation}

\noindent where
\begin{equation}
\begin{split}
\label{eq:gru_gates}
    \Tilde{\vh}_{t} &= \tanh(\mW^{(h)} \vx_{t} + \mU^{(h)} (\vr_{t} * \vh_{t-1}) + \vb^{(h)})  \\
    \vu_{t} &= \sigma(\mW^{(u)} \vx_{t} + \mU^{(u)} \vh_{t-1} + \vb^{(u)}) \\
    \vr_{t} &= \sigma(\mW^{(r)} \vx_{t} + \mU^{(r)} \vh_{t-1} + \vb^{(r)}).
\end{split}
\end{equation}

The number of parameters for either of the above RNN units is $g D (M + D)$, where $g$ is the number of distinct gates, which is 4 for an LSTM and 3 for a GRU. Given any factorization of the input and hidden dimensions into positive integers as $D = \prod_{k=1}^n d_k$ and $M = \prod_{k=1}^n m_k$ (where $d_k, m_k \geq 1$), this parameter count can be expressed as
\begin{align}
\label{eq:params_rnn}
    N_{dense} &= g D (M + D) = g \left(\prod_{k=1}^n d_k m_k + \prod_{k=1}^n d_k^2 \right) \nonumber \\
    &= \bigO(d^n (m^n + d^n)),
\end{align}

\noindent where $d = \max_k d_k$ and $m = \max_k m_k$. This version of the parameter count will allow for an easier comparison of typical RNN models with the tensorized RNNs introduced below.

\subsection{Tensor-Train Decomposition}
The tensor-train (TT) decomposition, introduced in~\cite{oseledets2011} and equivalent to the earlier matrix product state model of many-body physics~\cite{vidal2003}, gives a method for representing higher-order tensors as a type of iterated low-rank factorization. A TT representation of an $n$th-order tensor $\tT \in \R^{p_1 \times p_2 \times \cdots \times p_n}$ is a tuple of $n$ tensors $\tGn = (\tG_1,\tG_2,...,\tG_n)$, called the TT cores. Each core has dimension $\tG_k \in \R^{p_k \times r_{k-1} \times r_k}$, where the $r_k$ for $k \in \{1, \ldots, n-1\}$ are hyperparameters called the TT ranks of the model. Given a collection of TT cores, the tensor $\tT$ associated with these cores has elements given by the following vector-matrix-vector products
\begin{equation}
\begin{split}
\label{eqn:tt_format}
    \tT(i_1, i_2, \ldots, i_n) &= \tG_1(i_1) \tG_2(i_2) \cdots \tG_d(i_n),
    %&= \sum_{j_0, \cdots, j_d = 1}^{r_0, \cdots, r_d} \tG_1(i_1, j_0, j_1) \cdots \tG_d (i_d, j_{d-1}, j_d)
\end{split}
\end{equation}
where $\tG_k(i_k) = \tG_k(i_k, :, :) \in \R^{r_{k-1} \times r_k}$ indicates an index-dependent matrix associated with the $k$th core, with each $i_k \in [p_k]$ and $r_0, r_n$ each taken to be 1. We will refer to $\tT$ as the ``global'' tensor encoded by the TT cores, which constitute a ``local'' representation of $\tT$.

The TT decomposition is capable of exactly representing any $n$th-order tensor given sufficiently large TT ranks using the TT-SVD procedure of~\cite{oseledets2011}, but a more common practice is to fix the TT ranks at small values and use the core tensors as a compact parameterization which is optimized to minimize some loss function defined on the global tensor. This approach is not limited to cases where higher-order tensors are already present, as any vector $\vv$ with dimension $P = \prod_{k=1}^n p_k$ can be reshaped into an $n$th order tensor $\tV \in \R^{p_1 \times \cdots \times p_n}$. Such ``TT vectors'' provide an efficient description requiring only $\sum_{k=1}^n p_k r_{k-1} r_k = \bigO(\log(P))$ parameters when all TT ranks $r_k$ and core dimensions $p_k$ are bounded, compared with $P$ parameters for a dense representation.

The same procedure can be applied to matrices of shape $D \times M$ when $D = \prod_{k=1}^n d_k$ and $M = \prod_{k=1}^n m_k$, yielding a TT matrix defined by $n$ tensor cores. In this case we choose each TT core $\tG$ to have four indices with respective dimensions $d_k$, $m_k$, $r_{k-1}$, and $r_k$, and denote the associated index-dependent matrices by $\tG_k(i_k, j_k) = \tG_k(i_k, j_k, :, :) \in \R^{r_{k-1} \times r_k}$, for $i_k \in [d_k]$ and $j_k \in [m_k]$.

% \jake{} Something about key properties of the tt format (esp those relevant to this work): supports several arithmetic operations, efficient computational time, scales to arbitrary no of dimensions,...

\subsection{Tensorizing Neural Networks}
The bulk of the parameters in a neural network consist of large weight matrices represented in dense format. It was shown in~\cite{novikov2015} that the representation of these matrices as TT matrices allowed for a significant reduction in parameter count, while introducing little or no additional error in the performance of the network.

Given a weight matrix $\mW$ of shape $D \times M$, where $D = \prod_{k=1}^n d_k$ and $M = \prod_{k=1}^n m_k$, then the affine transformation implemented as part of a typical neural network layer takes the form $\vy = \mW \vx + \vb$. In a tensorized neural network, $\vx, \vy, \vb$ are represented normally as dense vectors, while the weight matrix $\mW$ is represented in TT form. The affine transformation is carried out by first using multilinear tensor contractions to perform the multiplication $\mW \vx$, with $\vx$ reshaped into a dense $n$th order tensor $\tX$, and then using standard dense addition for the bias vector $\vb$. The output vector $\vy$ can be described in reshaped form as the tensor $\tY$ with elements
\begin{multline}
\label{eq:tt_layer}
% \begin{split}
    \tY(i_1, \cdots\!, i_n) = \tB(i_1, \cdots\!, i_d)\ + \\
    \sum_{j_1, \cdots, j_n}\!\!\! \left(\tG_1(i_1,j_1) \cdots \tG_d(i_d,j_d)\right) \tX(j_1, \cdots\!, j_d).
% \end{split}
\end{multline}

% \begin{equation}
% \begin{split}
%     &\tY(j_1, \cdots, j_n) = \\ 
%     &\sum_{i_1, \cdots, i_d} \mathcal{W}((i_1,j_1), \cdots, (i_d, j_d))\tX(i_1, \cdots, i_d) \\
%     &+ \tB(j_1, \cdots, j_d)  
% \end{split}
% \end{equation}

By carrying out the above summations (including those implicit in the matrix-vector products) in an optimal order,~\eqref{eq:tt_layer} can be evaluated with a total cost of $\bigO(n r^2 d M)$, where $r = \max_k r_k$. In the typical setting where $r$, $m$, and $d$ remain bounded as $D$ and $M$ are increased, this cost is $\bigO(\log(\max(D, M)) M)$, compared to $\bigO(D M)$ for the usual affine map. This representation is also compact, requiring only $\bigO(n r^2 d m) = \bigO(\log(\max(D, M)))$ parameters, compared to $\bigO(D M)$ parameters for a dense representation.

% Moreover, this representation is compact. The total number of parameters in $\{\tG_k\}_{k=1}^n$ is $\sum_{k=1}^n d_k m_k r_{k-1} r_k$, that is the sum of element count of all cores. In comparison, the original weight matrix stores $\prod_{k=1}^n d_k m_k$ parameters. One can check that if the ranks $r_k$ are small, the tensor-train parameters will be much smaller than that of the dense weight matrix.

For clarity, we refer to a fully-connected layer represented in tensor-train form as a tensor-train layer (TTL), and denote the linear portion of the operation implemented in~\eqref{eq:tt_layer} as $\TTL(\vx; \tGn)$.

\subsection{Tensorizing RNNs}
\label{sec:tensorizing_rnns}
We describe a straightforward application of the above tensorization procedure to LSTM models, as utilized in~\cite{tjandra2017,yang2017tensor}, which allows for a significant reduction in the models' parameter count. In the next section we propose an extension of this procedure which permits an even greater degree of compression to be attained. %but our method of gate concatenation followed by tensorization can be applied more broadly to networks where there exist many linear or affine maps sharing common input and output dimensions, such as other gated neural network architectures.

% In order to fully tensorize the LSTM, we factorize the weight parameters of the gates as a tensor train. By doing this we essentially convert all fully connected layers into a TTL. The bias terms are absorbed into the TTLs for the hidden-hidden transformations.

An LSTM recurrent unit contains 8 weight matrices, each providing contributions to one of the four independent gate vectors coming from an input vector $\vx_{t}$ or previous hidden vector $\vh_{t}$. When these matrices are replaced by tensor-train matrices,~\eqref{eq:lstm_gates} can be re-written as
\begin{equation}
\begin{split}
\label{eq:vanilla_ttlstm}
    &\Tilde{\vc}_{t} = \tanh(\TTL(\vx_{t}; \tGn^{(Wc)}) + \TTL(\vh_{t-1}; \tGn^{(Uc)}) + \vb^{(c)})  \\
    &\vu_{t} = \sigma(\TTL(\vx_{t}; \tGn^{(Wu)}) + \TTL(\vh_{t-1}; \tGn^{(Uu)}) + \vb^{(u)}) \\
    &\vf_{t} = \sigma(\TTL(\vx_{t}; \tGn^{(Wf)}) + \TTL(\vh_{t-1}; \tGn^{(Uf)}) + \vb^{(f)}) \\
    &\vo_{t} = \sigma(\TTL(\vx_{t}; \tGn^{(Wo)}) + \TTL(\vh_{t-1}; \tGn^{(Uo)}) + \vb^{(o)}).
\end{split}
\end{equation}

Each of the 8 weight matrices $V_e$ (where $V$ is one of $W$ or $U$, and $e$ is one of $c$, $u$, $f$, or $o$) is replaced by its own collection of tensor-train cores $\tGn^{Ve}$, and we assume for simplicity that the same factorization of $D = \prod_{k=1}^n d_k$ and $M = \prod_{k=1}^n m_k$ is used for each of the 8 tensor-train matrices.

For a tensorized gated RNN with $g$ gates and an identical factorization for each tensor-train matrix, such as the LSTM above, the total parameter count is
\begin{align}
\label{eq:params_vanilla_ttrnn}
    N_{TT1} &= g \sum_{k=1}^n r_{k-1} r_k d_k (m_k + d_k) \nonumber \\
            &= \bigO(g n r^2 d (m + d)).
\end{align}

Although the exact comparison of this count to~\eqref{eq:params_rnn} depends on the TT ranks $r_k$ and the number of cores $n$ employed, it is clear that for the typical case where $r, d, m \ll \min(M, D)$, a tensorized RNN will require significantly fewer parameters. However, the use of a separate TT matrix for each gate in the RNN unit still leads to a multiplicative factor of $g$ in~\eqref{eq:params_vanilla_ttrnn}.

% It is easy to see that if $r_k$ is small, the number parameters in the tensorized format (Eq. \ref{eq:params_vanilla_ttrnn}) will always be much smaller than that of the regular RNN (Eq. \ref{eq:params_rnn}).

\section{Fully Tensorized RNNs}
We now introduce a different tensorization method, where a tensor-train factorization is applied to entire collections of concatenated weight matrices, rather than to individual matrices. The efficient nature of the tensor-train decomposition leads to a further reduction in model parameters, with LSTMs requiring approximately four times fewer parameters compared to the tensorization above. We show more generally that gated RNNs with $g$ gates exhibit a roughly $g$-fold reduction in the parameter count with this method, on top of the already sizable reduction coming from the use of tensor-train matrices.

\subsection{Gate Concatenation}
We achieve further compression of our tensorized RNN by jointly tensorizing the input-hidden weights, as well as the hidden-hidden weights. Taking the LSTM as an example, we first take the row-wise concatenation of the four input-hidden matrices $\mW^{(c)}, \mW^{(u)}, \mW^{(f)}, \mW^{(o)} \in \R^{D \times M}$, which gives a single input-hidden matrix $\mW \in \R^{4D \times M}$. More concretely, the concatenated weight matrices utilized are
%
%Notice that the number of TTLs in the tensorized RNN unit scales with the number of gates, i.e., $2\times g$. This could lead to a large number of individual core tensors in the unit. For instance, if each TTL has 4 cores, then there would be a total of 32 tensors. This could pose optimization difficulties for the model. We address this by 
%
\begin{equation}
\begin{split}
    &\mW = [ \mW^{(c)}, \mW^{(u)}, \mW^{(f)}, \mW^{(o)}]^T, \\
    &\mU = [ \mU^{(c)}, \mU^{(u)}, \mU^{(f)}, \mU^{(o)}]^T.
\end{split}
\end{equation}

For regular LSTMs with dense weight matrices, this concatenation gives a means of replacing four separate matrix-vector multiplications by a single larger multiplication, permitting greater parallelism. After the single vector $\mW \vx$ is computed it can be split into four equal-sized pieces, each holding the value of one of the gate vectors.

\begin{figure}[t]
    \centering
    \includegraphics[width=\columnwidth]{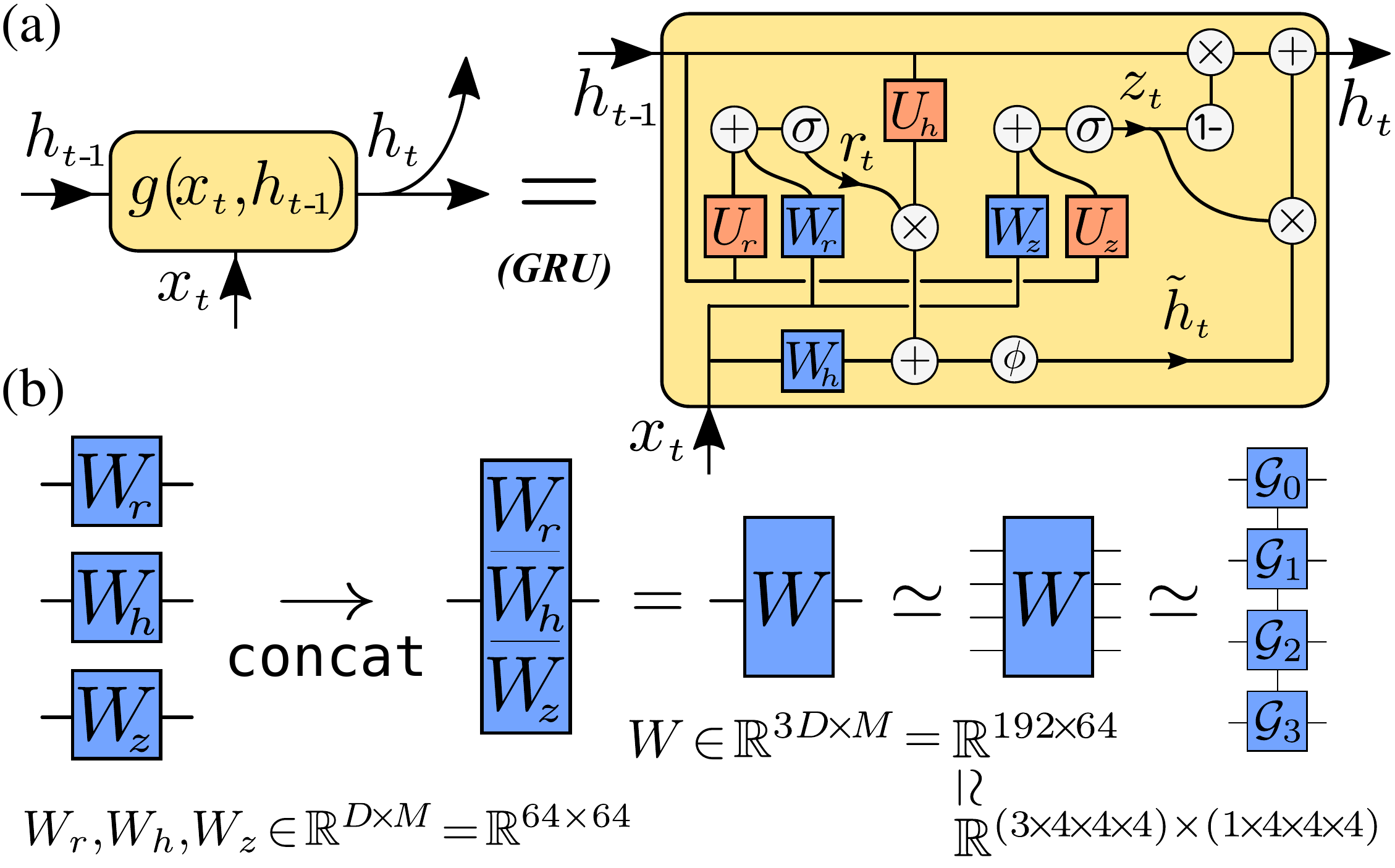}
    \caption{Illustrated of tensorization process on GRU cell with hidden and input dimensions $D = M = 64$.\\ \textbf{(a)} Layout of the recurrent update function $g(x_{t}, h_{t-1})$ for a GRU, with biases omitted for simplicity. Weight matrices are shown in blue and orange, with matrices of the same color having the same shape. In traditional RNNs, these weight matrices are parameterized as separate dense matrices. \textbf{(b)} Our compression process involves first concatenating all matrices of the same type, then tensorizing this composite matrix by parameterizing it as a TT matrix. For the given case, the stacked matrix $W \in \R^{192 \times 64}$ is represented as a tensor $\tT \in \R^{(3\times4\times4\times4) \times (1\times4\times4\times4)}$, which in turn is represented by the contraction of four TT cores $\tG_k \in \R^{d_k \times m_k \times r_{k-1} \times r_k}$. In the particular case shown, the bottom cores $\tG_k$ for $k = 1, 2, 3$ give a family of $r_0$ matrices $\mM_\alpha \in \R^{64 \times 64}$ jointly represented in TT format, while the top core $\tG_0$ acts as a matrix assigning each of the GRU gate matrices to a linear mixture of the TT matrices $\mM_\alpha$ (see Section~\ref{sec:weight_sharing}).
    }
    \label{fig:tensorized_rnn_cell}
\end{figure}

When the concatenated weight matrices are represented as a tensor-train layer, this leads to the revised LSTM gate equations,
\begin{equation}
\label{eq:ttlstm}
\begin{split}
    \Tilde{\vc}_{t} &= \tanh(\TTL(\vx_{t}; \tG^{W})_{1} + \TTL(\vh_{t-1}; \tG^{U})_{1} + \vb^{(c)})  \\
    \vu_{t} &= \sigma(\TTL(\vx_{t}; \tG^{W})_{2} + \TTL(\vh_{t-1}; \tG^{U})_{2} + \vb^{(u)}) \\
    \vf_{t} &= \sigma(\TTL(\vx_{t}; \tG^{W})_{3} + \TTL(\vh_{t-1}; \tG^{U})_{3} + \vb^{(f)}) \\
    \vo_{t} &= \sigma(\TTL(\vx_{t}; \tG^{W})_{4} + \TTL(\vh_{t-1}; \tG^{U})_{4} + \vb^{(o)}),
\end{split}
\end{equation}

\noindent where $\TTL(\vx_{t}; \tG^{W})_i$ and $\TTL(\vh_{t-1}; \tG^{U})_i$ are the $i$th equally-sized vectors in the TT matrix-vector products associated with $W$ and $U$, which contribute to the $c$, $u$, $f$, and $o$ gates. This process can be carried out analogously for a gated RNN with $g$ gates, where the matrices $\mW \in \R^{gD \times M}$, $\mU \in \R^{gD \times D}$ are each concatenations of $g$ separate matrices. An example of this process for a GRU model is given in Figure~\ref{fig:tensorized_rnn_cell}.

% While this process looks completely analogous to the dense case, we will see that this use of the tensor-train decomposition gives a significant further compression of the model.

\subsection{Compression and Runtime}

When tensorizing the individual weight matrices of an RNN in Section~\ref{sec:tensorizing_rnns}, the hidden and input dimensions were factored into $n$ smaller terms, as $D = \prod_{k=1}^n d_k$ and $M = \prod_{k=1}^n m_k$. For the case of concatenated weight matrices $\mW$ and $\mU$, a closely related factorization can be employed, namely $gD = \prod_{k=0}^n d_k$ and $M = \prod_{k=0}^n m_k$, where we take $d_0 = g$ and $m_0 = 1$, along with identical $d_k, m_k$ for all $k \geq 1$. 

Taking $\mW$ as an example, a tensor-train decomposition relative to this augmented factorization will give the collection of $n+1$ cores $\tGno^{(W)} = (\tG^{(W)}_0, \tG^{(W)}_1, \ldots, \tG^{(W)}_n)$, where the cores $\tG^{(W)}_k$ for $k > 1$ are shaped identically to a tensor-train factorization of any one of the single-gate weight matrices. The single new core appearing in this decomposition has a shape of $\tG^{(W)}_0 \in \R^{g\times 1\times 1\times r_0}$, for a new TT rank parameter $r_0$, and removing the singleton indices gives a matrix $\mV^{(W)} \in \R^{g \times r_0}$. This leads to a revised parameter count of
\begin{align}
\label{eq:params_ttrnn}
    N_{TT2} &= g r_0 + \sum_{k=1}^n r_{k-1} r_k d_k (m_k + d_k) \nonumber \\
            &= \bigO(n r^2 d (m + d)),
\end{align}

\noindent giving a compression ratio approximately $g$ times greater than~\eqref{eq:params_vanilla_ttrnn}. Using an example model in Table~\ref{tab:numerical_example}, we illustrate the level of compression and speedup in inference time that can be obtained for different configurations of our fully tensorized RNNs. This shows particular promise for the application of RNN models in settings with limited resources, such as edge devices. Finally, the training time for TT-RNNs is comparable to untensorized RNNs, although with a clear dependence on the TT rank.

\subsection{Weight Sharing}
\label{sec:weight_sharing}
Some intuition for this parameter reduction can be gained by interpreting the concatenated global matrix $\mW$ encoded by the TT cores $\tGno^{(W)}$ in terms of the small matrix $\mV^{(W)}$ coming from the first core $\tG^{(W)}_0$. Seen this way, the contraction of the remaining TT cores $(\tG^{(W)}_1, \ldots, \tG^{(W)}_n)$ gives a tensor which encodes a family of $r_0$ matrices $\{\mM_\alpha \in \R^{D \times M}\}_{\alpha=1}^{r_0}$. Contracting all of the TT cores (including $\tG^{(W)}_0$) and selecting the $i$th subspace then gives a single-gate weight matrix $\mW_i$, which corresponds to the linear mixture of matrices
\begin{equation}
    \mW_i = \sum_{\alpha=1}^{r_0} \mV^{(W)}_{i, \alpha} \mM_\alpha.
\end{equation}

Since all of the matrices $\mW_i$ are jointly encoded as a collection of $n$ TT cores whose matrix dimensions are identical to those of a single tensorized gate matrix, specifying the weight matrices for all $g$ gates in this manner requires a comparable number of parameters to specifying a single weight matrix in TT format.

\begin{table}[t]
\caption{Comparison of model size and per-step training and inference times of RNNs and TT-RNNs. Each model has a single recurrent layer with hidden size of 512, a linear projection layer of embedding size 256, and input dimension of 4,096. Each TT-RNNs has 2 cores, and $r$ denotes the TT rank. For both LSTM and GRU models, the tensorized versions achieve significant compression of model parameters while reducing the inference time and, for smaller values of $r$, decreasing the training time. All reported times were obtained on an Intel(R) Xeon(R) CPU E5-1650 v3 @ 3.50GHz with 128GB of RAM, and averaged over 100 runs.} \smallskip
\label{tab:numerical_example}
\centering
\resizebox{.98\columnwidth}{!}{
\smallskip\begin{tabular}{|c|crrr|}
\hline
Model & $r$ & \# params & Train time (s) & Eval. time (s) \\
 \hline
LSTM & $-$ & 9,570,560 & $12.84 \pm .17$ & $3.70 \pm .19$  \\
% \cline{2-5}
\hline
& 2 & 21,248 & $9.37 \pm .11$ & $2.13 \pm .13$ \\
TT-LSTM & 3 & 30,720 & $11.92 \pm .22$  & $2.23 \pm .13$ \\
& 4 & 40,192 & $15.55 \pm .37$  & $2.48 \pm .25$  \\
\hline
GRU & $-$ & 7,212,288 & $10.12 \pm .26$ &  $2.53 \pm .07$ \\
% \cline{2-5}
\hline
& 2 & 19,200 & $8.09 \pm .21$ & $1.43 \pm .09$ \\
TT-GRU & 3 & 27,136 & $9.18 \pm .15$  & $1.59 \pm .08$  \\
& 4 & 35,072 & $11.23 \pm .30$ & $1.80 \pm .10$  \\
 \hline
\end{tabular}}
\end{table}

% \begin{table}[t]
% \caption{An example comparing model size and evaluation time of RNNs and TT-RNNs. The model has single recurrent layer with a hidden size of 512 and a single linear projection layer with embedding size 256. $r$ is the TT rank. The TT-RNNs have 2 cores shaped $16 \times 32$ for the hidden units.} \smallskip
% \label{tab:numerical_example}
% \centering
% \resizebox{1\columnwidth}{!}{
% \smallskip\begin{tabular}{|c|c|crrr|}
% \hline
% Input size & Model & $r$ & \# params & Compr. & Eval. time (s) \\
%  \hline
% %  \hline
% \multirow{4}{*}{256} & LSTM &  & 1,706,240 & & $0.12 \pm .001$ \\
%   \cline{2-6}
%  &  & 2 & 12,032 &   & $0.16 \pm .001$ \\
%  & TT-LSTM & 3 & 16,896 &   & $0.16 \pm .001$ \\
%  &  & 4 & 21,760 &   &  $0.17 \pm .001$\\
  
%   \hline
% \multirow{4}{*}{4096} & LSTM &  & 9,570,560 &  & $0.24 \pm .005$  \\
%   \cline{2-6}
%   &  & 2 & 21,248 & & $0.17 \pm .003$ \\
%   & TT-LSTM & 3 & 30,720 &   & $0.17 \pm .002$ \\
%   &  & 4 & 40,192 &   & $0.18 \pm .002$  \\
%  \hline
% \end{tabular}}
% \end{table}

% \subsubsection{Smoother Optimization} 
% The gate concatenation trick guarantees that every TT-RNN unit has only 2 TTLs, instead of $2 \times g$, thereby further constraining the total number of individual core tensors in the model. We evaluate the optimization stability of TT-RNNs in our experiments by visualizing the distributions of activations and gradients. To do this we pool the local activations and gradients of the core tensors to global formats using:

\section{Experiments}
We benchmark the performance of TT-RNN models using experiments on image classification and speaker verification tasks. Results for TT-LSTM are reported here, while those for TT-GRU can be found in the supplementary material. Beyond assessing the accuracy in these tasks, we characterize trade-offs between compression and accuracy arising from different choices of TT rank and core layout. In the process, we find that the tensor-train parameterization acts as a form of regularization, leading to improved stability and generalization during training.

For simplicity and ease of comparison, all models in the following are trained without explicit regularization such as dropout, weight decay, or gradient clipping. The tensorized models were written in PyTorch~\cite{paszke2019} using the tensor-train implementation from~\cite{khrulkov2019}, and are available on GitHub\footnote{https://github.com/onucharles/tensorized-rnn}.

\subsection{Permuted Pixel MNIST}
We first evaluate the TT-LSTMs on the permuted sequential MNIST task~\cite{lecun1998mnist} in which the $28 \times 28$ pixel images of handwritten digits are randomly rearranged using a fixed permutation into sequences of length 784. These are split into 50k training, 10k validation, and 10k test images, with the validation dataset used to determine the end of training by early stopping. 

The LSTM and TT-LSTM were each chosen as single-layer models with 256 hidden units. Training was performed with a batch size of 256 and Adam optimizer, using a piecewise constant learning rate starting at 0.001.% (except for the TT-RNNs, which were stable under a faster training of 0.01). 

Table~\ref{tab:permuted-mnist-perf} reports the digit classification accuracy, where the hidden dimensions of the TT-LSTM are factored into either 2 or 3 TT cores using TT ranks of 2, 4, or 6 to connect adjacent cores. Although a clear tradeoff is present between compression and accuracy, even the largest TT-LSTM utilizes 46 times fewer parameter in total, while achieving comparable performance to the LSTM baseline ($-0.28\%$ classification accuracy).

\begin{table}[h]
\caption{Comparison of TT-RNN and standard RNN models on the permuted pixel MNIST task. The models use a single-layer containing $D = 256$ hidden units, and are trained identically. The performance of the TT-RNNs varies with the parameter count, but achieves comparable accuracy to a standard RNNs while maintaining a compression ratio of 46 times and 25 times fewer parameters, in the TT-LSTM and TT-GRU respectively.}\smallskip
\label{tab:permuted-mnist-perf}
\centering
\resizebox{.95\columnwidth}{!}{
\smallskip\begin{tabular}{|c|c|crrr|}
\hline
Model   & Cores & $r$ & \#Params & Compr. & Acc. (\%) \\
\hline
LSTM    & $-$   & $-$ & 266,762 & $-$ & 89.77     \\
\cline{1-6}
\multirow{6}{*}{TT-LSTM} & \multirow{3}{*}{2} & 2   &  3,434 &  78  &  87.98    \\
 &     & 4   & 5,834   &  46  &   89.49   \\
 &  & 6  &  8,234  &  32  &   89.22  \\
\cline{2-6}
 & \multirow{3}{*}{3} & 2   & 1,842   & 145   &   85.36   \\
 &     & 4   &  3,354  &  80  &  87.18   \\
 &  & 6  &  5,570  &  48  & 89.30    \\
\hline
GRU    & $-$   & $-$ & 201,482 & $-$ & 91.49     \\
\cline{1-6}
\multirow{6}{*}{TT-GRU} & \multirow{3}{*}{2} & 2   &  3,674 &  55  &  87.94    \\
 &     & 4   & 5,802   &  35  &   89.29   \\
 &  & 6  &  7,930  &  25  &   90.26  \\
\cline{2-6}
 & \multirow{3}{*}{3} & 2   & 2,282   &  88  &   87.62   \\
 &     & 4   &  3,722  &  54  &  88.90   \\
 &  & 6  & 5,866  &  34  & 89.80    \\
\hline
\end{tabular}}
\end{table}

\subsection{Speaker Verification}
In the speaker verification problem, the objective is to ascertain if an utterance of speech belongs to a given individual, based on a collection of utterances labeled by individuals. We use the LibriSpeech dataset, containing around 1,000 hours of English language audiobook recordings~\cite{panayotov2015librispeech}, where training, validation, and testing are carried out on the {\em train-clean-100}, {\em dev-clean}, and {\em test-clean} partitions. 

The TT-LSTM exhibits impressive performance in this more complex task, substantially improving on the LSTM baseline while using 200 times fewer parameters. The tensor-train parameterization of the TT-LSTM appears to represent a form of implicit regularization, which leads to less overfitting while also minimizing the issue of vanishing and exploding gradients during training.

\subsubsection{Setup}
Our model for speaker verification contains two main components, an utterance encoder and a similarity function, as in~\cite{heigold2016end,xie2019utterance}. The utterance encoder consists of an RNN which computes fixed-dimensional embeddings from spectograms of input utterances, while the similarity function assigns similarity scores to pairs of embeddings. %The loss function used to train the network is designed such that embeddings of utterances of the same speaker are close to each other, while those of different speakers are far apart. 

We use the generalized end-to-end (GE2E) loss function~\cite{wan2018generalized} to train the model, which encourages embeddings of utterances to cluster based on the associated speaker. Given an embedding vector $\ve_{ji}$ for the $i$th utterance by the $j$th speaker, the GE2E loss is
\begin{equation}
\begin{split}
    L(\ve_{ji}) &= -\mS_{ji,j} + \log \sum_{k=1}^N \exp(\mS_{ji,k}), \\
\end{split}
\end{equation}
where $\mS_{ji,j} = w \cdot \cos(\ve_{ji}, \vc_k) + b$ is the scaled cosine similarity between the embedding $\ve_{ji}$ and the centroid of the embeddings of speaker $j$, denoted $\vc_j$. The scaling coefficients $w$ and $b$ are initialized to $10$ and $-5$ respectively. The full loss is then the sum of all utterance-specific losses, $L = \sum_{j,i} L(\ve_{ji})$. %See~\cite{wan2018generalized} for more details.

We report performance in the speaker verification task using the equal error rate (EER) metric, which is the error rate on the receiver-operating characteristic (ROC) curve when the false positive rate and false negative rates are equal.

\begin{figure}[!b]
\centering
\includegraphics[width=.95\columnwidth]{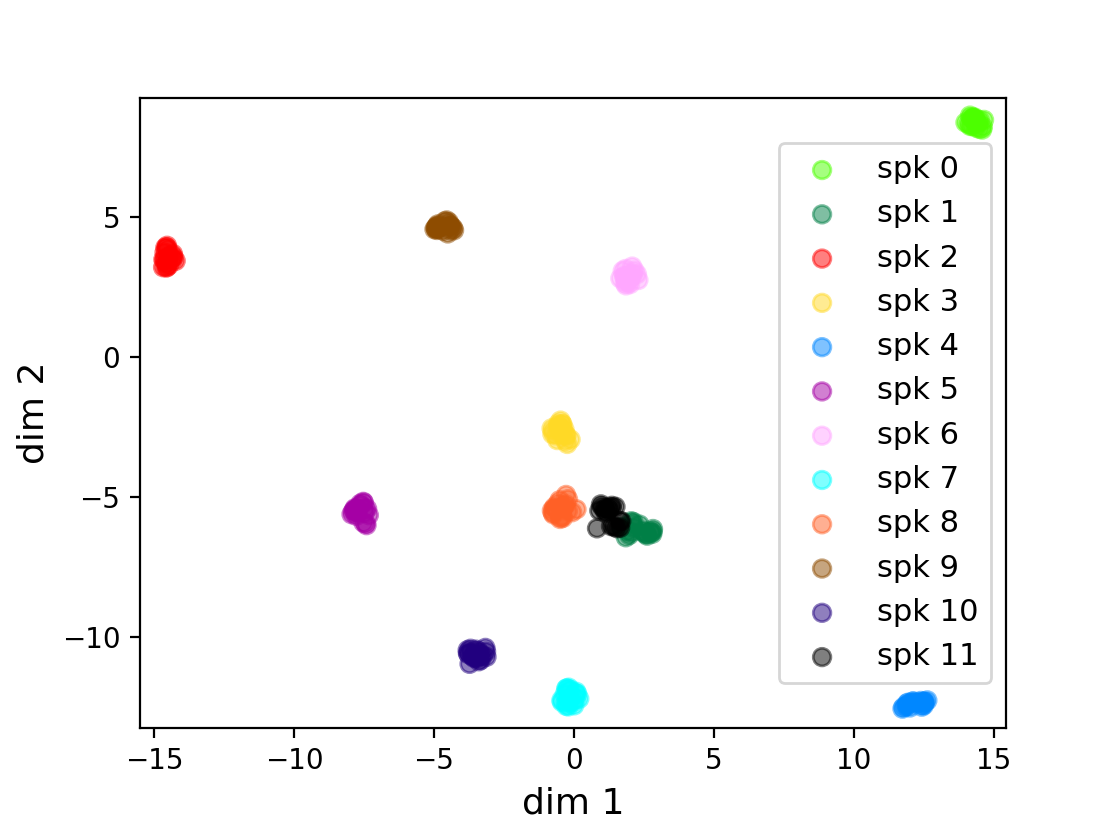}
\caption{Low dimensional UMAP visualization of embeddings from the TT-LSTM. Each datapoint corresponds to the 256-dimensional embedding of an utterance, where colors reflect the identity of different speakers. A clear clustering pattern is seen amongst the utterances from each speaker.}
\label{fig:umap_embeds}
\end{figure}

\begin{figure*}[h]
\centering
\includegraphics[width=2\columnwidth]{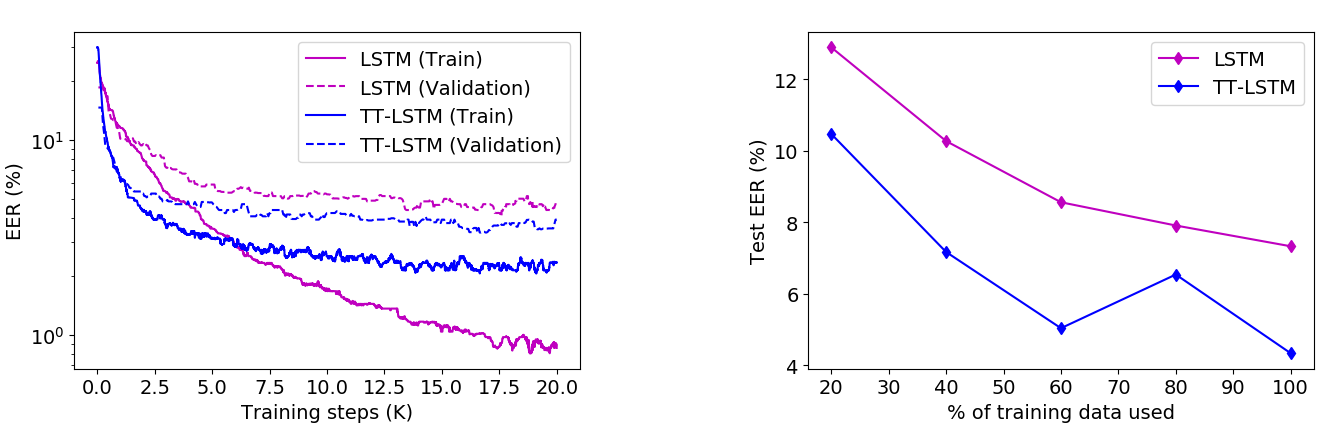}
\caption{Illustration of regularization benefits of TT-RNNs. \textbf{Left}: Learning curves for the best LSTM and TT-LSTM models. The use of tensor-train weights acts as an implicit regularizer, raising the training error while reducing the margin between training and validation EERs of the TT-LSTM compared to the LSTM. \textbf{Right}: Performance of the models using different fractions of the LibriSpeech training set. Each datapoint gives the test EER of the corresponding model after training, and we see the TT-LSTM consistently generalizing better than the standard LSTM.}
\label{fig:regularisation}
\end{figure*}

\begin{figure}[!b]
\centering
\includegraphics[width=1\columnwidth]{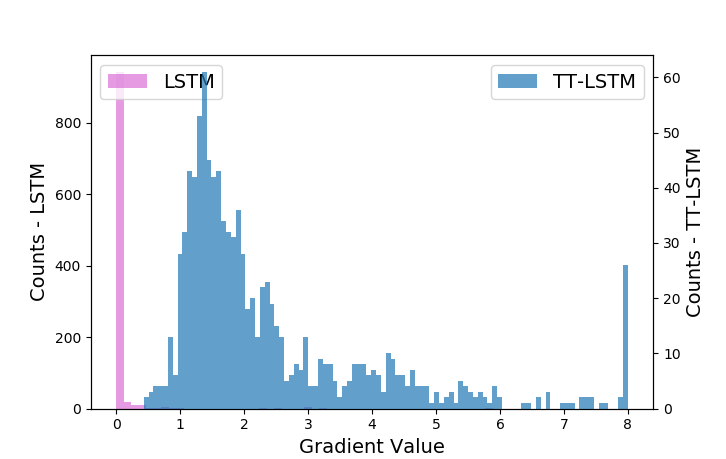}
\caption{Distribution of the norm of gradients of model parameters across 1000 training steps. Model is same configuration as before, but learning rate is increased to 0.01 from 0.001. The LSTM succumbs to the vanishing gradient problem, while the gradients of the TT-LSTM remain distributed over a wide range.}
\label{fig:grad_hist}
\end{figure}

\subsubsection{Performance}
Our utterance encoder consists of a single-layer LSTM with hidden size of 768, whose output is converted to an embedding of dimension 256 using a fully-connected linear layer. The input to this encoder is 40-bin $\times$ 160-frame Mel spectograms of utterances. We compare regular LSTMs and TT-LSTMs for these identical input, hidden, and embedding dimensions, as given in Table~\ref{tab:speaker_verif_perf}.

%and has .  We compare this with our tensorized model TT-LSTM which has the same hidden and embedding dimensions. We train all configurations of TT-LSTM with number of cores, $d = \{ 2, 3, 4\}$ and ranks $r = \{ 1, 2, 3, 4, 5\}$.  
%The projection layer is also tensorized (as a TTL) using the same values of $d$ and $r$.

Using a standard LSTM in the encoder gives an EER of 7.33\%, similar to the performance found in~\cite{zhou2019training}. By contrast a TT-LSTM encoder led to significantly better EERs, with the best configuration achieving an EER of 4.34\%. This increased accuracy was accompanied by a reduction in the total parameter count, from 2.6M parameters to only 13K. By reducing the TT rank, this parameter count can be further reduced while still maintaining higher accuracy than the LSTM baseline.

%The smallest TT-LSTM has only 4K parameters compared to over 2.6M in the LSTM and achieves a better EER of 6.09\%.

\begin{table}[t]
\caption{Performance of RNNs and TT-RNNs on the task of speaker verification. Models have a single layer with 768 hidden units and a linear projection layer of 256. The lowest ranked TT-RNNs outperform the RNNs on this more challenging task of speaker verification, achieving larger compression ratios of 653 (TT-LSTM) and 369 (TT-GRU). \cut{The TT-RNNs have a hidden shape of ($16 \times 16$) when 2 cores and ($4 \times 8 \times 8$) when 3 cores.} EER is the equal error rate (lower is better).}\smallskip
\label{tab:speaker_verif_perf}
\centering
\resizebox{.97\columnwidth}{!}{
\smallskip\begin{tabular}{|c|c|crrr|}
\hline
Model   & Cores & $r$ & \#Params & Compr. & EER (\%) \\
\hline
LSTM    & $-$   & $-$ & 2,682,114 & $-$   & 7.33     \\
\cline{1-6}
\multirow{6}{*}{TT-LSTM} & \multirow{3}{*}{2} & 1   &  8,178 &  328  &  4.71    \\
 &      & 2   & 13,026  &  206  &  4.34    \\
 &     & 4   &  22,722  & 118   & 6.21  \\
\cline{2-6}
 & \multirow{3}{*}{3} & 1   &  4,106 & 653   &  6.09    \\
 &      & 2   & 5,394  &  497  & 5.31     \\
 &     & 4   &  9,506  &  282  & 5.38     \\
\hline
GRU    & $-$   & $-$ & 2,063,106 & $-$   & 7.87     \\
\cline{1-6}
\multirow{6}{*}{TT-GRU} & \multirow{3}{*}{2} & 1   &  9,074 &  227  &  5.31    \\
 &      & 2   & 13,282  & 155   &  6.72    \\
 &     & 4   &  21,698  &  95  & 5.36  \\
\cline{2-6}
 & \multirow{3}{*}{3} & 1   &  5,594 & 369   &  6.46    \\
 &      & 2   & 6,738  &  306  & 6.39     \\
 &     & 4   &  10,274  &  201  & 4.48     \\
\hline
\end{tabular}}
\end{table}

Analyzing the embeddings learned by the TT-LSTM further demonstrates the performance of the model in speaker verification. We use uniform manifold approximation and projection (UMAP)~\cite{mcinnes2018umap} to project the 256-dimension embedding vectors into 2D space (Figure~\ref{fig:umap_embeds}), which shows that the embeddings learned by the TT-LSTM effectively cluster the utterances from each speaker. 

\subsubsection{Regularization}
%The TT-LSTMs were far easier to train, requiring minimal to no hyperparameter tuning. We suspect that the performance gain comes from this as well as its implicit regularization. 
TT-LSTMs utilize a more compact set of weight parameters, which can be expressed as a low-dimensional family of weight matrices. To assess if this low-dimensional parameterization has benefits for regularization, we first examine the learning curves of TT-LSTMs and standard LSTMs during training (Figure~\ref{fig:regularisation}, left). We observe that while LSTM encoders achieve lower loss during training, this loss is not reflected in the validation loss, likely due to overfitting. By contrast, the TT-LSTM shows better generalization, giving a smaller discrepancy between training and validation loss, and ultimately a lower validation EER.

To further test this generalization, we conduct the speaker verification experiments in a more data-limited setting, using between 20\% and 100\% of the training data. TT-LSTMs consistently performed better than the LSTM baseline when trained with small amounts of data (Figure~\ref{fig:regularisation}, right).

\subsubsection{Training Stability}
 We observed during the initial hyperparameter search an increased robustness in the performance of TT-LSTMs relative to changes in the learning rate. Both LSTMs and TT-LSTM models were trained at a learning rate of 0.001, but increasing this to 0.01 led to an instability in the former and no noticeable impact on the latter. The distribution of gradients for this case is given in Figure~\ref{fig:grad_hist}. The standard LSTM exhibits vanishing gradients, effectively saturating at 0, while the gradients for TT-LSTM are distributed over a reasonable range. 

\section{Related Work}
The compression of deep neural networks (DNNs) has been of interest for a long time. It has been shown that DNNs are typically parameterized in a redundant fashion, allowing the prediction of values of some parameters of a trained model given knowledge of the others~\cite{denil2013}.

Several approaches depend on some kind of post-processing after a large model has been trained. Model distillation~\cite{ba2014,hinton2015} for example is a successful technique which retrains a smaller model by using the output activations of the trained large model as labels, instead of the actual data labels. This was found to result in smaller models that are fast to train and match the performance of the larger models from which they were distilled. Quantization is another post-processing technique which uses a more coarse-grained representation for each parameter value, thereby reducing the memory needed to store a trained model's parameters. One complication with these post-processing methods is that they are not end-to-end; the process of pruning the DNN is separate from training.

Matrix and tensor factorization techniques provide an alternative that is end-to-end trainable. A natural first step is to decompose parameter weight matrices in a low-rank matrix factorized format. This was done in~\cite{sainath2013} to compress the last fully-connected layer of a convolutional neural network (CNN). Restricting to the last layer is limited in the compression achieved, since the other layers of the network themselves contain many parameters. However, utilizing this approach in internal layers results in a lower effective number of hidden units, ultimately hurting accuracy.

Tensor factorization methods, such as that employed here, can generally be used to decompose matrices in higher-dimension space, and were used in~\cite{yu2017} to capture higher order interactions in dynamical processes. The idea of tensorizing neural networks in an end-to-end trainable manner using tensor-train decomposition was first introduced in~\cite{novikov2015}, where a fully-connected layer was reshaped and factorized as a tensor in TT format to achieve impressive compression. The extension to convolutional layers was later given in~\cite{garipov2016ultimate}.

Different aspects of these ideas were extended to recurrent neural networks (RNN) in~\cite{yang2017tensor,tjandra2017}. The work of~\cite{yang2017tensor} applied tensor-train layers to the large encoding matrices used for high-dimensional video input, allowing for simultaneous compression and improved performance in video classification. This was later followed by~\cite{yin2020}, which reported further gains through the use of the more complex Hierarchical Tucker decomposition in place of tensor trains.

By contrast, \cite{tjandra2017} tensorized RNNs by assigning a separate TT matrix to each of the separate weight matrices in a recurrent cell, with a focus on GRU models. This allowed significant compression to be achieved not only with high-dimensional inputs, but also with high-dimensional hidden states. Our work is similar to~\cite{tjandra2017}, but achieves further compression by jointly tensorizing the weights within each RNN cell. We show how this process leads to a novel form of weight sharing, which is verified experimentally to have tangible benefits for performance and compression. The use of a tensor-train parameterization is shown to represent an implicit regularization capable of improving training and generalization. Our TT-RNN model is available as open-source code, and can be used as a drop-in replacement for standard RNN models.

\section{Acknowledgement}
Onu's research is supported by a Vanier Canada Graduate Scholarship. Precup's work is supported by the Canadian Institute for Advanced Research (CIFAR).
\cut{
\section{Appendices}
Provide training details, hyperparameters, etc for all 3 experiments.
}

\bigskip
\newpage

\bibliography{main}

\end{document}

%% file: math_commands.tex
\DeclareMathAlphabet\mathbfcal{OMS}{cmsy}{b}{n}

\DeclareMathOperator{\TTL}{TTL}
\newcommand{\tGn}{\tG_{[n]}}
\newcommand{\tGno}{\tG_{[n+1]}}

\def\1{\bm{1}}
\newcommand{\bigO}{\mathcal{O}}

\def\va{\textbf{\textit{a}}}
\def\vb{\textbf{\textit{b}}}
\def\vc{\textbf{\textit{c}}}

\def\ve{\textbf{\textit{e}}}
\def\vf{\textbf{\textit{f}}}

\def\vh{\textbf{\textit{h}}}

\def\vo{\textbf{\textit{o}}}

\def\vr{\textbf{\textit{r}}}

\def\vu{\textbf{\textit{u}}}
\def\vv{\textbf{\textit{v}}}

\def\vx{\textbf{\textit{x}}}
\def\vy{\textbf{\textit{y}}}

\def\mM{\textbf{\textit{M}}}

\def\mS{\textbf{\textit{S}}}

\def\mU{\textbf{\textit{U}}}
\def\mV{\textbf{\textit{V}}}
\def\mW{\textbf{\textit{W}}}

\DeclareMathAlphabet{\mathsfit}{\encodingdefault}{\sfdefault}{m}{sl}
\SetMathAlphabet{\mathsfit}{bold}{\encodingdefault}{\sfdefault}{bx}{n}
\newcommand{\tens}[1]{\mathbfcal{#1}}

\def\tB{{\tens{B}}}

\def\tG{{\tens{G}}}

\def\tT{{\tens{T}}}

\def\tV{{\tens{V}}}

\def\tX{{\tens{X}}}
\def\tY{{\tens{Y}}}

\newcommand{\R}{\mathbb{R}}